# Efficient Off-Policy Reinforcement Learning via Brain-Inspired Computing

Yang Ni
yni3@uci.edu
University of California, Irvine
Irvine, California, USA

Danny Abraham
dannya1@uci.edu
University of California, Irvine
Irvine, California, USA

Mariam Issa
mariamai@uci.edu
University of California, Irvine
Irvine, California, USA

Yeseong Kim
yeseongkim@dgist.ac.kr
Daegu Gyeongbuk Institute of Science and Technology
Daegu, Republic of Korea

Pietro Mercati
pietro.mercati@intel.com
Intel Labs
Hilsboro, Oregon, USA

Mohsen Imani
m.imani@uci.edu
University of California, Irvine
Irvine, California, USA

## ABSTRACT

Reinforcement Learning (RL) has opened up new opportunities to enhance existing smart systems that generally include a complex decision-making process. However, modern RL algorithms, e.g., Deep Q-Networks (DQN), are based on deep neural networks, resulting in high computational costs. In this paper, we propose QHD, an off-policy value-based Hyperdimensional Reinforcement Learning, that mimics brain properties toward robust and real-time learning. QHD relies on a lightweight brain-inspired model to learn an optimal policy in an unknown environment. On both desktop and power-limited embedded platforms, QHD achieves significantly better overall efficiency than DQN while providing higher or comparable rewards. QHD is also suitable for highly-efficient reinforcement learning with great potential for online and real-time learning. Our solution supports a small experience replay batch size that provides 12.3× speedup compared to DQN while ensuring minimal quality loss. Our evaluation shows QHD capability for real-time learning, providing 34.6× speedup and significantly better quality of learning than DQN.

## 1 INTRODUCTION

Smart systems and services generally reside in a highly-dynamic yet unknown environment and require intelligent algorithms to make optimal decisions with little prior knowledge. In recent years, Reinforcement Learning (RL) has opened up new opportunities to solve a wide range of complex predictions and decision-making tasks that were previously out of reach for a machine [1]. Compared to supervised and unsupervised learning methods, RL does not have direct access to labeled training data. Learning through agent-environment interaction makes RL appealing to dynamic control and automated system optimization such as smart transportation and smart grid, where the optimal policy is hard to define and is constantly changing with its environment [1, 2].

RL methods are generally categorized into policy-based and value-based RL. The policy-based method directly parameterizes the policy and optimizes it via on-policy model training. On the other hand, value-based RL supports off-policy training, i.e., all past interactions can be used toward learning. Therefore, value-based RL is much more sample-efficient. As one of the most popular value-based methods, Deep Q-Networks (DQN) exploits DNN to learn an approximation of the Q-value for every pair of actions and state. Recently, there have been active developments for various DQN applications such as playing computer games [3], Genomics [4], and smart city [2, 5]. DQN is capable of learning complex tasks without modeling the environment, but its power comes at a price, i.e., the huge computation cost and long learning time. This makes it only suitable for powerful computers in the cloud. However, offloading RL to the cloud not only leads to extra communication overhead but also causes security and privacy concerns.

Therefore, we redesign the RL algorithm by exploiting the brain-inspired highly-efficient HyperDimensional Computing (HDC) [6]. HDC is motivated by how human brains process different kinds of inputs, i.e., brains express information using a vast number of neurons. The information is then processed and memorized in a holistic and high-dimensional way. For inputs in the lower-dimensional space, HDC encodes them to vectors of several thousand dimensions, i.e., hypervectors. The learning process is based on highly-parallelizable operations of hypervectors. HDC has been applied as a lightweight machine learning solution to multiple applications where it is capable of achieving comparable accuracy to DNN with significantly higher efficiency [7, 8].

However, current HDC solutions mainly focus on traditional classification and clustering. In contrast, in this paper, we propose QHD, a value-based Hyperdimensional Reinforcement Learning algorithm with off-policy training, which mimics brain properties towards robust and real-time learning. The main contributions of the paper are listed as follows:

- To the best of our knowledge, QHD is the first off-policy value-based hyperdimensional RL algorithm targeting discrete action space. QHD relies on lightweight HDC models to learn an optimal policy in an unknown environment. Our algorithm maps state-action space into high-dimensional space for efficient decision-making via novel brain-inspired HDC encoding and self-learning.
- Thanks to the brain-like hyperdimensional operations, QHD can utilize even a small amount of available training data. It thereby supports a much smaller training batch and experience buffer than DQN while still providing high-quality results.
- We compare our QHD accuracy and efficiency with the DQN algorithm for multiple dynamic control tasks. Our evaluation shows that QHD achieves significantly better overall efficiency than DQN, especially on the power-limited embedded platform, e.g., up to about 15× speedup. For real-time learning, QHD provides 34.6× speedup and significantly better quality.

## 2 RELATED WORK

**Reinforcement Learning:** In recent years, RL algorithms have obtained dramatically more attention because of the advancement in deep learning. For example, DQN greatly expands the application of RL to fields like computer games [3, 9], transportation optimization [2, 10], and health care [11, 12]. In [10], researchers focus on driver dispatch optimization within online ride-sharing services. They use DQN to learn a policy for matching available drivers and users to maximize the success rate while minimizing the wait time. All works above utilize DNN to handle complex agent-environment interactions, so they are computationally intensive with insufficient efficiency. In contrast, we propose a brain-inspired reinforcement learning solution with inherent efficiency and robustness.

**Hyperdimensional Computing:** Prior HDC works mainly provide solutions to classification and cognitive tasks, such as graph reasoning [13], bio-signal processing [7, 8], speech recognition [14], neuromorphic sensing [15] and multi-sensor signal classification [16]. In these highlighted applications, HDC has outperformed state-of-the-art learning solutions, e.g., support vector machines [14] and neural networks [16]. Recent orthogonal work proposes an HDC-based policy-based RL specifically for continuous control tasks [17]. However, this work does not provide support for RL tasks with discrete action space. In addition, as mentioned in Section 1, policy-based RL methods are less sample-efficient due to the lack of off-policy training. Unlike all prior works, this paper is the first effort focusing on hyperdimensional off-policy value-based RL.

## 3 QHD: HYPERDIMENSIONAL Q-LEARNING

### 3.1 Overview

Fig. 1 shows an overview of QHD supporting hyperdimensional reinforcement learning. In our RL task, there are two components (Agent and Environment) and three variables (Action, State, and Reward). Fig. 1(a) exploits a Cartpole example to illustrate these components and variables. As shown in Fig. 1, the interaction between the agent and environment forms a loop in which the action taken based on the current state leads to the next state and reward. The trajectory of each episode is saved in local memory for later learning. In Fig. 1(b), we provide an overview of QHD algorithm guiding the agent in the decision-making process.

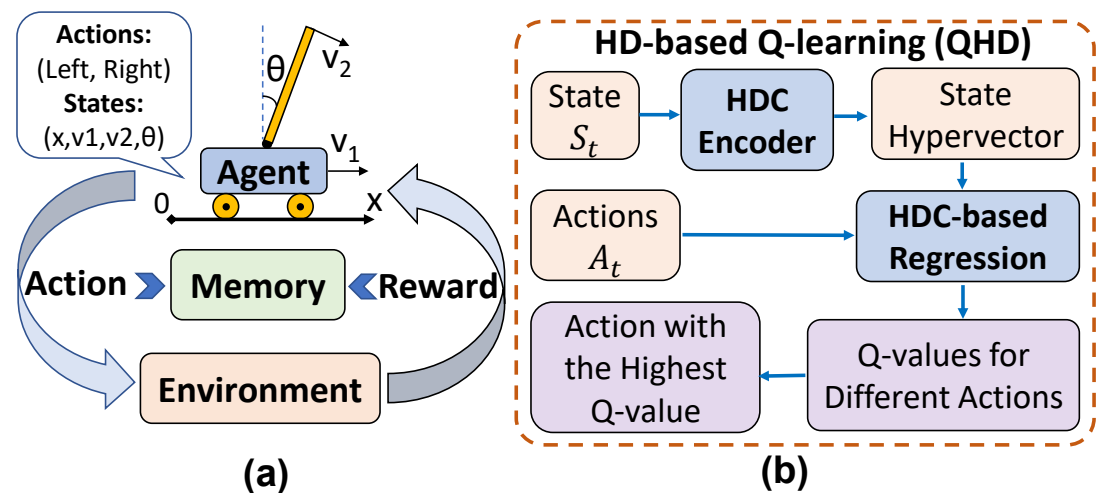

**Figure 1: Overview of QHD reinforcement learning.**

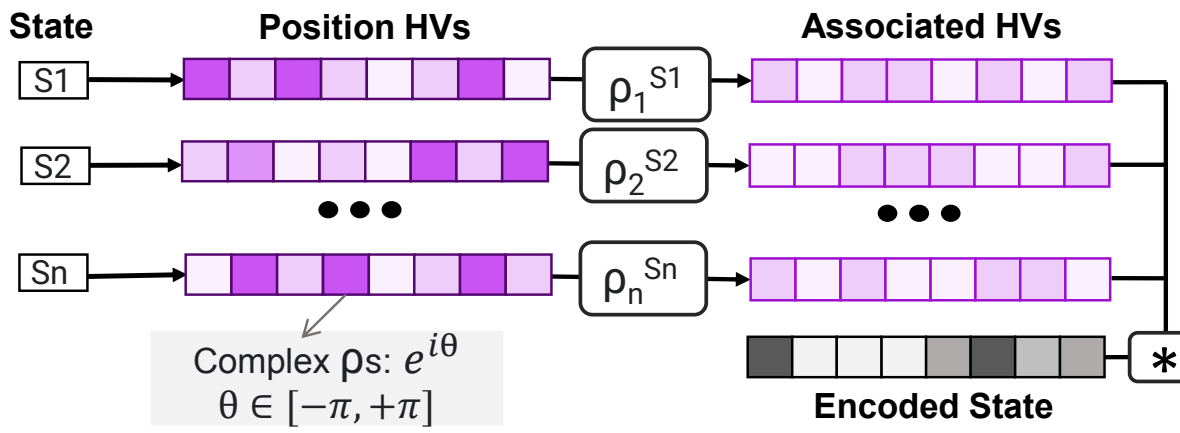

**Figure 2: HDC encoding with complex-valued position hypervectors (HVs).**

### 3.2 QHD Hyperdimensional Encoding

QHD starts by mapping the current state vector from the original to high-dimensional space, i.e., hypervector encoding. Notice that we can create a large number of near-orthogonal hypervectors through random sampling, i.e., their dot product $\vec{\rho}_1 \cdot \vec{\rho}_2 \approx 0$. Our solution encodes inputs using hypervectors with random exponential elements, e.g., $\vec{\rho}_1$ belongs to $\{e^{i\theta} : \theta \in [-\pi, \pi]\}^D$. $D$ is the dimensionality of these hypervectors.

As Fig. 2 shows, through well-defined hypervector operations, the original information is evenly distributed across all hypervector elements, i.e., a holographic representation. The advantage of being holographic is that we can accumulate information by simply combining two hypervectors. The complex-valued position hypervectors used in this paper enable the encoder to capture the correlation between input features in finer granularity.

Next, we define the following HDC mathematics that manipulates input information in high-dimensional space:

**Continuous Binding**: The goal of binding is to associate items in hyperspace. Assuming we have an $n$-element state vector at time step $t$ in the RL tasks: $S_t = \{s_1, s_2, \ldots, s_n\}$. Our encoding generates a random exponential hypervector for each state element $\{\vec{\rho}_1, \vec{\rho}_2, \cdots, \vec{\rho}_n\}$ and then associates state elements with these hypervectors: $\vec{\mathcal{S}}_t = \vec{\rho}_1^{\,s_1} * \vec{\rho}_2^{\,s_2} * \cdots * \vec{\rho}_n^{\,s_n}$. We define $\vec{\rho}_k^{\,s_k}$ to be the component-wise exponential of $\vec{\rho}_k$.

**Bundling:** This operation stands for component-wise addition of hypervectors. The bundling operation is the core of memorization for HDC models, in which the information from multiple hypervectors is saved into one single hypervector. The bundled hypervector is similar to every component hypervector, i.e., $(\vec{\rho}_1 + \vec{\rho}_2) \cdot \vec{\rho}_1 >> 0$. Thus, we represent sets using the bundling operation. Bundling of several encoded states results in a model hypervector: $\vec{\mathcal{M}} = \vec{\mathcal{S}}_1 + \vec{\mathcal{S}}_2 + \cdots + \vec{\mathcal{S}}_m$. In Section 3.3, we leverage a weighted bundling to represent it in high-dimensional space, i.e., $\vec{\mathcal{M}} = \alpha_1\vec{\mathcal{S}}_1 + \alpha_2\vec{\mathcal{S}}_2 + \cdots + \alpha_m\vec{\mathcal{S}}_m$. The $\alpha$s are learned via HDC-based regression.

**Hypervector Similarity:** Our HDC encoder aims at preserving the distance relationship among inputs. It maps similar input state vectors to similar locations in the high-dimensional space, i.e., the similarity between encoded hypervectors is close to 1. To verify this, we define a hypervector similarity metric: $\delta(\vec{\rho}^{\ x} \cdot \vec{\rho}^{\ y}) = \vec{\rho}^{\ x} \cdot \vec{\rho}^{\ \dagger y}/D$, where $\vec{\rho}^{\ \dagger y}$ is the element-wise conjugate. Then, assuming $x \approx y$, we have: $\vec{\rho}^{\ x} \cdot \vec{\rho}^{\ \dagger y}/D = D^{-1}\sum_d e^{i\theta_d(y-x)} \approx 1$.

## 3.3 QHD Hyperdimensional Regression

We develop a regression model based on hyperdimensional mathematics. Our regression consists of multiple model hypervectors $\{\vec{\mathcal{M}}_1, \vec{\mathcal{M}}_2, \ldots, \vec{\mathcal{M}}_n\}$, where $n$ is the size of the action space. For evaluation of each action at time step $t$, we only select one of the model hypervectors $\vec{\mathcal{M}}_A$ that corresponds to a certain action $A$, and the regression is operated on the current-step state hypervector $\vec{\mathcal{S}}_t$. These model hypervectors are initialized to all zero elements and have the same dimensionality as the encoded state hypervector, i.e., $\vec{\mathcal{M}}_A \in \{0\}^{\mathcal{D}}$. In regression, the true value is given by the ideal Q-function, and we use hyperdimensional regression to approximately calculate the Q-value. We explain the ground truth for Q-value in Section 3.4 when we introduce QHD. On the other hand, the approximated Q-value for action $A$ equals the real component in the dot product between the model hypervector and the encoded state hypervector: $q_{pred} = real(\vec{\mathcal{M}}_A \cdot \vec{\mathcal{S}}^{\ \dagger}/D)$, where $\vec{\mathcal{S}}^{\ \dagger}$ is a conjugate of the encoded query with complex-valued elements (recall Section 3.2). As for the regression model update, we use the error between $q_{pred}$ and $q_{true}$ (ground truth). We either add or subtract a portion of the state hypervector to the model, weighted by the regression error: $\vec{\mathcal{M}}_A = \vec{\mathcal{M}}_A + \beta(q_{true} - q_{pred}) \times \vec{\mathcal{S}}$. This equation ensures that the model gets updated more aggressively for higher prediction error rates ($q_{true} - q_{pred} \gg 0$). The lightweight operations in our regression design, such as component-wise addition, contribute to the fast learning process for QHD.

## 3.4 Hyperdimensional Value-based Reinforcement Learning

In this section, we present the details for our QHD, a hyperdimensional Q-learning algorithm. We start our introduction with how agents with QHD make decisions at each time step. In QHD, we use a greedy policy that prefers actions with higher Q-values. However, it is crucial to balance the exploration of the environment and the exploitation of the learned model. We combine a random exploration strategy with the greedy policy, i.e., $\epsilon$-decay policy. Assuming the action space $\mathcal{A}$ and time step $t$:

$$A_t = \begin{cases} \text{random action } A \in \mathcal{A}, & \text{with probability } \epsilon \\ argmax_{A\in\mathcal{A}} Q(S_t, A), & \text{with probability } 1-\epsilon \end{cases} \quad (1)$$

The probability of selecting random actions will gradually drop after the agent explores and learns for several episodes. In experiments, we use a rate of changing $\epsilon$-decay less than 1; this ensures that QHD agents start to trust their learned model more while gradually lessening the importance of exploration. In the equation above, $Q(S_t, A)$ is a hyperdimensional regression model that returns approximated Q-values for input action-state pairs. Once an action $A_t$ is chosen by QHD, the agent interacts with the environment. We then obtain the new state $S_{t+1}$ for the agent and the feedback reward $R_t$ from the environment. At the next time step $t+1$, QHD selects another action $A_{t+1}$. This chain of actions and feedbacks form a trajectory or an episode until some termination conditions are met. To train an RL algorithm, these episodes or past experiences are usually saved to local memory as training samples. More specifically, we save a tuple of four elements for each step in the experience replay buffer: $(S_t, A_t, R_t, S_{t+1})$.

In DQN, the RL training process and parameter update are based on DNN back-propagation, while the training in QHD utilizes more efficient hypervector operations. The regression model in QHD is trained at the end of each time step after saving current information to the replay buffer. As in the DQN training process, we apply a strategy called experience replay. The experience replays in QHD samples a training batch and uses it for the regression model update. The training batch includes multiple tuples of past experiences.

Now assume we sample a one-step experience tuple from past trajectories to train our QHD, i.e., $(S_t, A_t, R_t, S_{t+1})$. In Section 3.3, we introduce the regression model update based on the approximation error. We first encode the input state $S_t$ to the hypervector $\vec{\mathcal{S}}_t$ and the predicted value $q_{t_pred}$ is simply calculated as:

$$q_{t_pred} = Q(S_t, A_t) = real(\vec{\mathcal{M}}_{A_t} \cdot \vec{\mathcal{S}}_t^{\ \dagger}/D) \quad (2)$$

For regression training, we need a ground truth $q_{t_true}$. We cannot directly obtain the true Q-value because RL is not typical supervised learning. For time step $t$, the feedback is the one-step reward $R_t$ while the Q-value is the expectation of accumulated rewards. The method to connect these two values is called the Bellman Equation or Dynamic Programming Equation [18]. Since most RL tasks can be viewed as a Markov Decision Process (MDP), the Bellman equation gives a recursive expression for the Q-value at step $t$, the expected sum of current rewards, and the Q-value for step $t+1$. To learn an optimal Q-function, we use the Bellman optimality equation as shown below:

$$q_{t_true} = R_t + \gamma max_A Q'(S_{t+1}, A) \quad (3)$$

Recall that our objective in QHD is to achieve optimal policy and maximize the accumulated rewards within one episode. The Bellman optimality equation states that to achieve optimal results for the whole task, we need to optimize each sub-task. Thus, the true value $q_{t_true}$ is the sum of $R_t$ and the max next-step Q-value. Instead of using model $Q$ to calculate the maximum next-step Q-value, we use a delayed model $Q'$ which gets updated periodically using parameters in $Q$. This method is called Double Q-learning [19]; it stabilizes the learning process and avoids the overestimation of Q-value caused by the maximization in the Bellman equation. We also include a reward decay term $\gamma$ that adjusts the effect of future rewards on the current step Q-value.

After obtaining the predicted Q-value and true Q-value, we perform regression model updates. We update the model corresponding to the action taken, using the regression error $q_{t_true} - q_{t_pred}$ and the encoded state hypervector. The learning rate is $\beta$.

$$\vec{\mathcal{M}}_{A_{t+1}} = \vec{\mathcal{M}}_{A_t} + \beta(q_{t_true} - q_{t_pred}) \times \vec{\mathcal{S}}_t \quad (4)$$

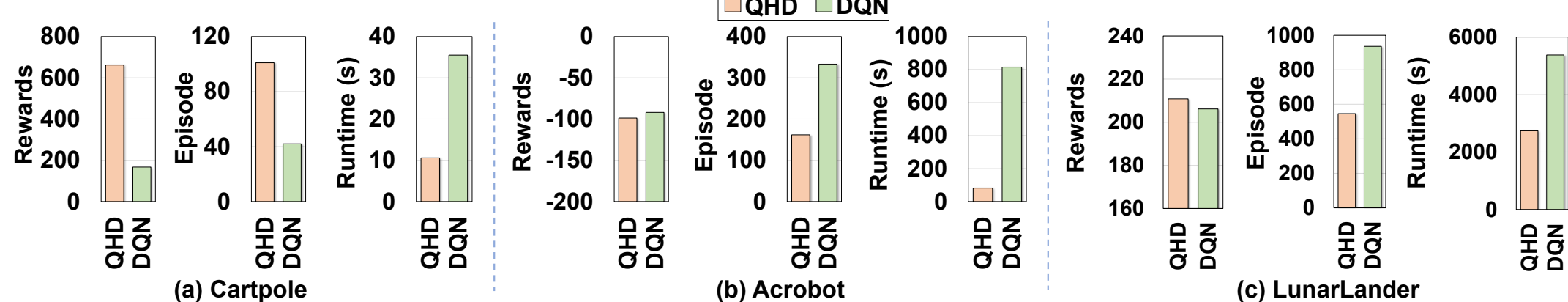

**Figure 3: Final reward and goal-achieved runtime for DQN and QHD.**

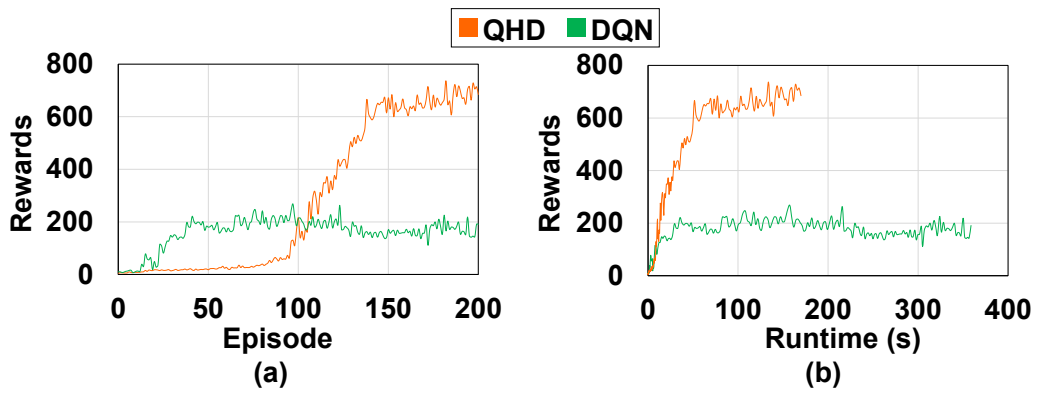

**Figure 4: QHD and DQN Cartpole rewards comparison: (a) is plot with episode index and (b) is plot with runtime as x-axis.**

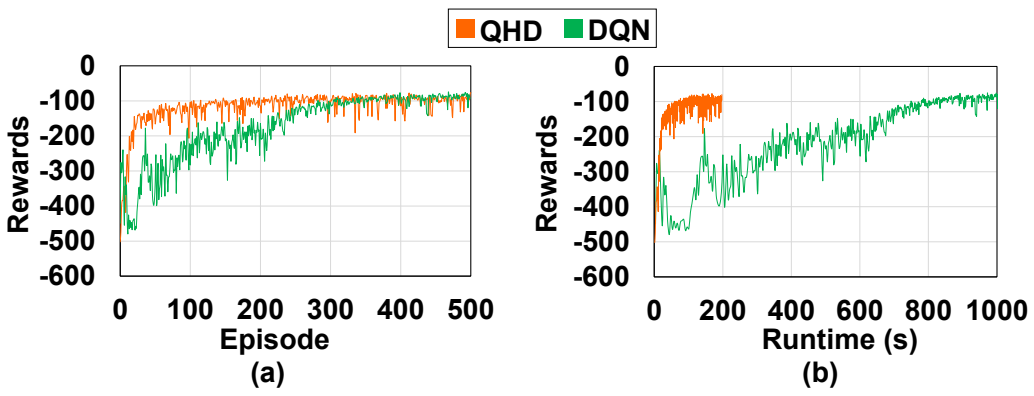

**Figure 5: QHD and DQN Acrobot rewards comparison with both episode index and runtime index.**

| Task | Acrobot | | Cartpole | | LunarLander | |
|---|---|---|---|---|---|---|
| Algorithm | DQN | QHD | DQN | QHD | DQN | QHD |
| Desktop CPU $T_{goal}$ (s) | 815 | 82 | 36 | 11 | 5379 | 2740 |
| RaspberryPi $T_{goal}$ (s) | 7041 | 476 | 435 | 72 | 51532 | 14483 |
| Scale Ratio | 8.6 | 5.8 | 12.3 | 6.8 | 9.6 | 5.3 |

**Table 1: QHD and DQN goal-achieved runtime ($T_{goal}$) comparison on both desktop and embedded platforms**

# 4 EXPERIMENTAL RESULT

## 4.1 Experiment Settings

We implement our QHD algorithm using Python on both desktop (Intel Core-i7 10700 with 65W TDP) and embedded hardware platforms (RaspberryPi 4 with 6W TDP). We validate the functionality of QHD with multiple control tasks in the OpenAI Gym [9]. For comparison, we use the DQN algorithm for the same tasks in our evaluation. In the following subsections, we compare these two methods' learning performance and efficiency in all tasks.

The regression model we used in QHD has dimensionality $D = 6000$ unless stated otherwise. This dimensionality setting provides us with a balance between learning quality and runtime cost; a larger dimensionality will generally lead to higher rewards achieved in the RL tasks with the cost of larger computation. The DQN is powered by a neural network with two hidden layers. The first layer has 128 neurons, and the second one has 256; except in the LunarLander task, where we use 64 neurons for the first layer and 128 for the second layer. The experience replay is enabled for model training in both methods, and we assume nearly unlimited replay buffer capacity for rewards and runtime comparison. We select different parameters for sampling training batches to ensure the best learning quality for both methods. The QHD training batch size is 4 for Acrobot/Cartpole and 10 for LunarLander, and the DQN training batch size is 64 for all tasks. Rewards and runtime results for both methods are averaged over multiple trials.

## 4.2 RL Rewards & Runtime comparison

Fig. 3 compares the performance of DQN and QHD over three popular OpenAI control tasks. As shown in Fig. 3(a) and 4, QHD achieves significantly higher final rewards in Cartpole compared to DQN. Within 200 episodes, QHD provides an averaged episodic reward over 660, which is nearly 4× higher than DQN. In the early episodes, this may lead to smaller rewards; but after the warm-up, QHD can quickly learn from accumulated experience and surpass DQN. Notice that considering the total execution time (shown in Fig. 4b), QHD is significantly faster than DQN and reaches higher rewards within the same amount of time. We also present the result for Acrobot in Fig. 3(b) and 5. Our QHD provides notably better learning efficiency compared to DQN, e.g., about 10× faster in terms of runtime and 2× fewer number of episodes. For LunarLander, we compare the RL performance and runtime in Fig. 3(c). It shows that compared to DQN, our QHD achieves the goal nearly 400 episodes earlier than DQN and about 2600 seconds (2 times) faster in runtime.

In Table 1, we collect the results for the implementation of QHD and DQN on embedded CPU. We show that the efficiency benefit of our algorithm remains significant in a power-limited environment. Thanks to the lightweight QHD learning process and the hypervector representation, our algorithm scales better than the deep network structure in DQN, i.e., a constantly smaller runtime scale ratio. In addition, the speedup brought by QHD is about 15× in Acrobot, 6× in Cartpole, and 3.55× in LunarLander.

## 4.3 Evaluate the effect of training batch size

Both QHD and DQN rely on experience replay, and since the experience replay buffer is ideally infinite, we need to sample the training dataset from the large replay buffer with a preset batch size. This parameter is rather crucial because it controls how much past experience is available for the agent to learn from, thereby deeply influencing the learning quality. A larger batch size prevents the agent from forgetting past experiences while bringing greater costs. Our QHD, on the other hand, aims to fully utilize the provided training samples at each step. In Fig. 6, we explore the effect of replay batch size on both methods. Fig. 6a compares the average rewards for the last 100 episodes, and it is clear that our QHD performs significantly better. For example, when the batch size is 2, QHD can still achieve the goal with an average of -102.9 rewards. On the other hand, DQN performs poorly with a reward of -480.9. This means that DQN does not efficiently utilize the limited available training samples.

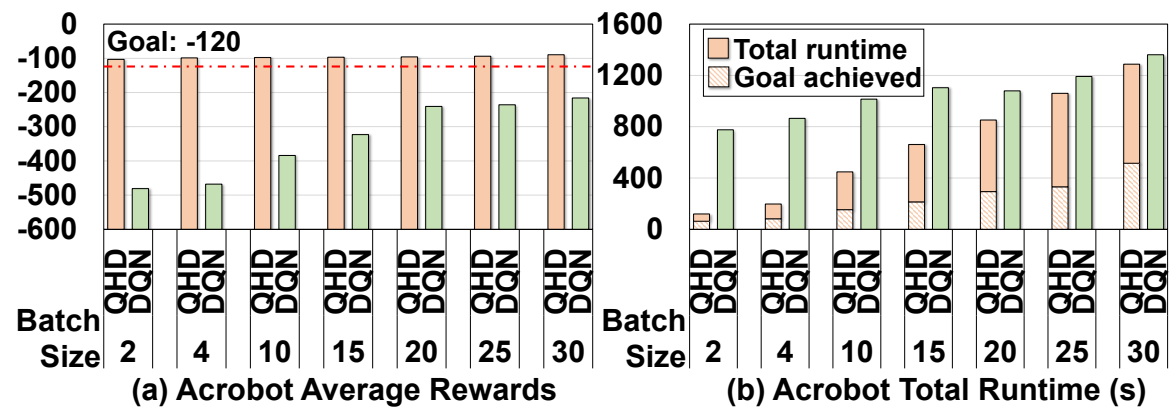

**Figure 6: Explore the effect of batch size.**

Apart from better performance, our QHD also provides higher efficiency. In Fig. 6(b), we provide both the QHD runtime for 500 episodes and when the goal is achieved. For DQN, only total runtime is provided because DQN cannot achieve the goal with small batch sizes from 2 to 30. Our QHD is constantly faster: with the batch size of 2 (15), our QHD is about 6.5× (1.7×) faster than DQN in terms of total runtime. Focusing on the actual runtime when achieving the target, QHD shows an even larger improvement, e.g., the speedup is about 12.3× (2.6×) with a batch size of 2 (30).

## 4.4 QHD vs. DQN with Limited Size of Experience Replay Buffer

In the above sections, we set the RL experience replay to have infinite capacity, i.e., the agent has access to all previous experiences during the training. However, in practical implementations, the memory available for experience replay is limited due to energy and space budgets. Thus, in this section, we evaluate the performance of our QHD with a tighter cap on the maximum replay buffer size and compare it to the DQN results.

In Fig. 7(a), we present the average reward achieved by both methods under different buffer sizes. The reward is averaged over the last 100 episodes. When collecting these results, we fix the training batch size; the batch size is 4 for QHD and 64 for DQN. The figure shows that DQN performs poorly when the buffer size is 64 and 128, with an average reward of -500. However, our QHD can reach that goal even with a buffer size as large as its batch size. These results show that QHD can perform RL tasks with online learning, i.e., a tiny replay buffer.

We also take one step further to explore the QHD capability of real-time learning. We set both the batch and buffer sizes to 1, which means the agent will learn based on only the current sample. We use DQN with a 256 buffer size and 64 batch size as an online-learning comparison. As shown in Fig. 7(b), with a larger buffer and batch size, DQN achieves significantly lower rewards (-345.4). For a 500-episode training, our QHD achieves average rewards of -113.7 using 83 seconds, which leads to a 34.6× speedup in total runtime.

## 5 CONCLUSION

We propose a novel lightweight value-based off-policy RL algorithm based on brain-inspired HDC. QHD utilizes HDC for high-quality

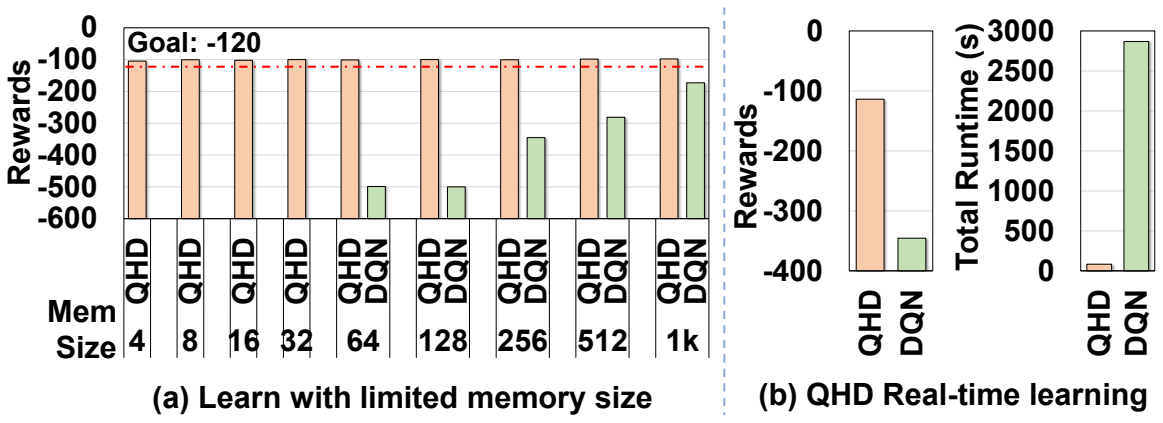

**Figure 7: QHD learning efficiency with tiny replay buffer.**

Q-value approximation and self-learning agent training. Our evaluation of several tasks shows that QHD provides significantly better efficiency and learning quality than DQN.

## ACKNOWLEDGEMENTS

This work was supported in part by National Science Foundation #2127780, Semiconductor Research Corporation (SRC), Department of the Navy, Office of Naval Research, grant #N00014-21-1-2225 and #N00014-22-1-2067, Air Force Office of Scientific Research, grant #FA9550-22-1-0253, Institute of Information & communications Technology Planning & Evaluation (IITP) grant funded by the Korea government(MSIT) (No.2022-0-00991, 1T-1C DRAM Array Based High-Bandwidth, Ultra-High Efficiency Processing-in-Memory Accelerator), and a generous gift from Cisco.